\documentclass{article}
\usepackage{spconf}
\usepackage{amsmath,amssymb,epsfig}
\usepackage{array}
\usepackage{booktabs}
\usepackage{graphicx}
\usepackage{epstopdf}
\usepackage{times}
\usepackage{color}
\usepackage{spconf,amsmath,graphicx}
\usepackage{threeparttable}
\usepackage{cite}
\usepackage {setspace}

\usepackage{spconf,amsmath,graphicx}
\usepackage{booktabs}
\usepackage{graphicx}                                                           
\usepackage{float}
\usepackage{amsfonts}
\usepackage{stfloats}
\usepackage{bbm}
\usepackage[colorlinks,linkcolor=red,anchorcolor=blue,citecolor=green]{hyperref}

\begin{document}
\title{MvCo-DoT: Multi-View Contrastive Domain Transfer Network for Medical Report Generation}

%

\name{\begin{tabular}{c}Ruizhi Wang$^{1}$, Xiangtao Wang$^{1}$, Zhenghua Xu$^{1,*}$\thanks{$^{*}$Corresponding author: zhenghua.xu@hebut.edu.cn (Zhenghua Xu).}, Wenting Xu$^{1}$, Junyang Chen$^{2}$, Thomas Lukasiewicz$^{3,4}$\end{tabular}}

\address{$^1$State Key Laboratory of Reliability and Intelligence of Electrical Equipment, \\School of Health Sciences and Biomedical Engineering, Hebei University of Technology, Tianjin, China\\
$^2$College of Computer Science and Software Engineering and Guangdong Laboratory
of Artificial\\ Intelligence and Digital Economy (SZ), Shenzhen University, Shenzhen, China\\
$^3$Institute of Logic and Computation, TU Wien, Vienna, Austria\\
$^4$Department of Computer Science, University of Oxford, Oxford, United Kingdom}



%
%

%
%
\maketitle
\begin{abstract}
 In clinical scenarios, multiple medical images with different views are usually generated at the same time, and they have high semantic consistency. However, the existing medical report generation methods cannot exploit the rich multi-view mutual information of medical images. Therefore, in this work, we propose the first multi-view medical report generation model, called MvCo-DoT. Specifically, MvCo-DoT first propose a multi-view contrastive learning (MvCo) strategy to help the deep reinforcement learning based model utilize the consistency of multi-view inputs for better model learning. Then, to close the performance gaps of using multi-view and single-view inputs, a domain transfer network is further proposed to ensure MvCo-DoT achieve almost the same performance as multi-view inputs using only single-view inputs.
 Extensive experiments on the IU X-Ray public dataset show that MvCo-DoT outperforms the SOTA medical report generation baselines in all metrics.

\end{abstract}
\vspace{-0.5em}
\begin{keywords}
Multi-view contrastive learning, Domain transfer, Medical report generation, Chest X-Ray
\end{keywords}

\vspace{-0.7em}
\section{Introduction}
\label{sec:intro}
\vspace{-0.5em}

Medical report generation is a multimodal cross-task in computer vision and natural language processing, which aims to reduce the workload of doctors by automatically generating diagnostic descriptions from medical images. Motivated by the application of deep learning in medical image analysis ~\cite{xu2019semi, 10.3389/fbioe.2023.1049555}, current medical report generation approaches typically utilize encoder-decoder architecture to learn medical images from different views but independently. And many spatial and channel attention methods~\cite{w-Net2022} have been carefully designed to explore multimodal interactions between image-level and sentence-level semantic features for report generation~\cite{anderson2018bottom,
xue2018multimodal,pan2020x,xu2020reinforced,hou2021ratchet}, however, this way of understanding images is not ideal for complex cross-modal generation tasks. Then, ~\cite{liu2021exploring} mitigates textual and visual data bias by exploring prior knowledge and posterior knowledge, ~\cite{chen2020generating} models and memorizes reports similar patterns between them, and thus promote Transformer to generate more informative long text interpretation reports. In order to utilize visual information more effectively, inspired by contrastive learning in the domain of natural images and natural language process~\cite{chen2020simple,he2020momentum,tian2020contrastive,gao2021simcse}, people started to try this method to improve the visual representation of medical images~\cite{yan2021weakly,zhang2022contrastive,
ZHANG-MIA2022}. However, the existing medical report generation methods have a common shortcoming that their inputs are usually single view,which fails to utilize the rich multi-view information in chest X-Ray images.

\begin{figure}[!t]
\centering
\vspace{-1em}
\includegraphics[scale=0.335]{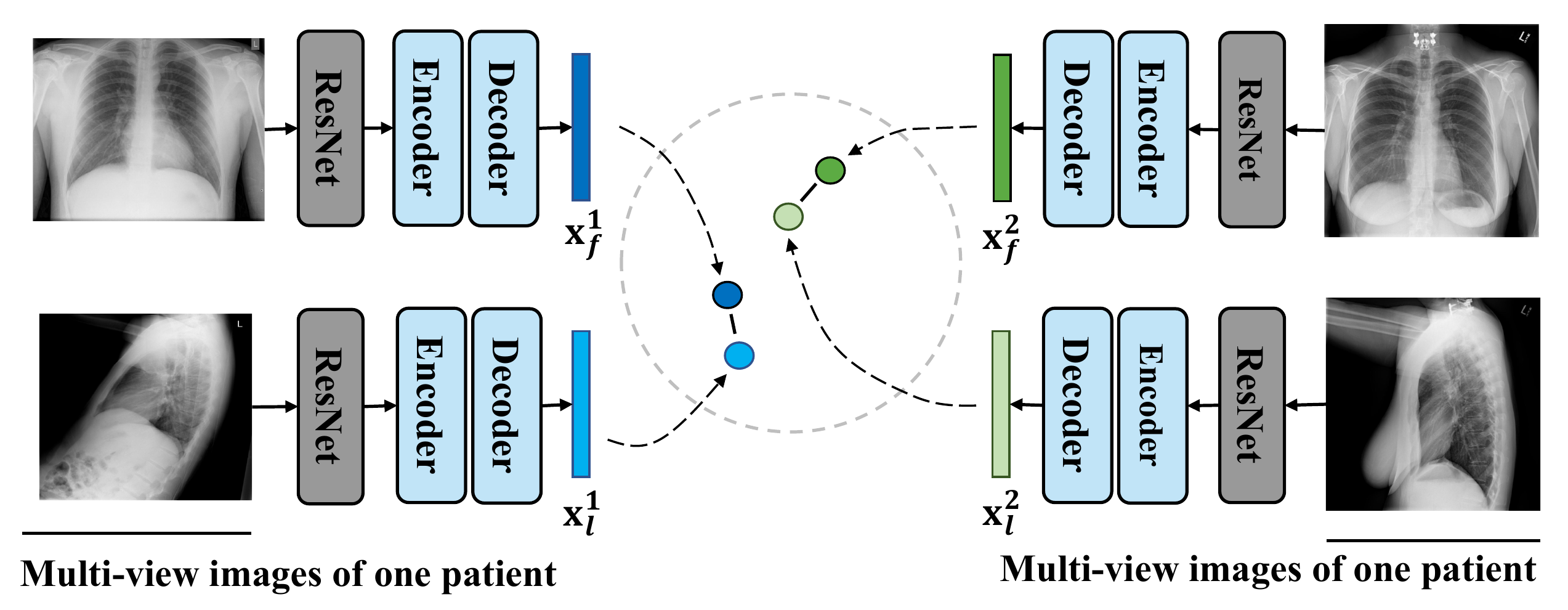}  
\vspace{-2.2em}
\caption{Semantic-based multi-view contrastive learning.
}  
\vspace{-1.85em}
\label{contrast}    
\end{figure}

Consequently, in this work, we propose a novel \textbf{M}ulti-\textbf{v}iew \textbf{Co}ntrastive
\textbf{Do}main \textbf{T}ransfer (\textbf{MvCo-DoT}) network for better medical report generation. Specifically, we first propose to integrate a Multi-view Contrastive Learning (\textbf{MvCo}) strategy into our previous deep reinforcement learning based medical report generation model~\cite{xu2020reinforced}. Intuitively, as shown in Fig.~\ref {contrast}, since the paired multi-view medical images are different imaging results of the same patient, their descriptions of lesions or organs in the patient should have high consistency~\cite{vu2021medaug}. Therefore, MvCo is proposed to utilize semantic embeddings of different views of patients' X-Ray images for contrastive learning. Compared to existing self-supervision based solutions~\cite{yan2021weakly,zhang2022contrastive}
whose contrastive learning modules are applied in encoders, feature representations used in MvCo is located in decoders, which thus have more direct impact on the quality of  resulting medical reports.
\begin{figure*}[!t]
\vspace{-2em}
\centering
\includegraphics[width=0.9\textwidth]{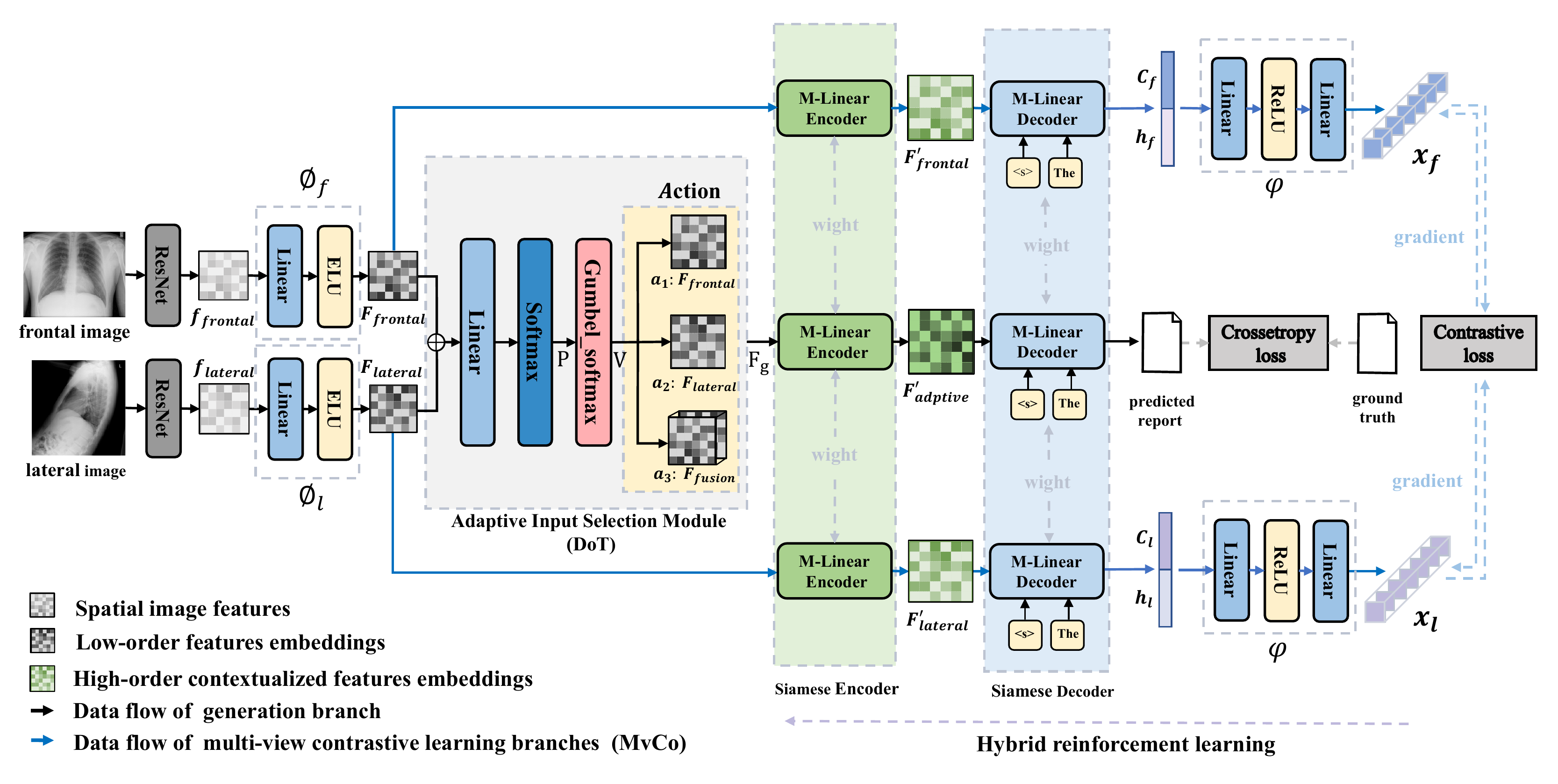}  
\vspace{-1.5em}
\caption{The architecture of our proposed MvCo-DoT network.}  
\vspace{-1.5em}
\label{architecture}    
\end{figure*}

In addition, our experimental studies show that although using MvCo can greatly improve the performances of medical report generation, it suffers from the problem of domain shift. When we have only single view X-Ray images as inputs, the inference results are greatly degraded because distributions of single-view inputs are very different from that of multi-view inputs. Consequently, we further propose to incorporate a Domain Transfer Network (\textbf{DoT}) into our medical report generation model to resolve this problem by closing the performance gaps of multi-view and single-view inputs.
Specifically, DoT is achieved by using a sampling-based adaptive input selection module, which enables generation branch to randomly select single or multi-view fused features as final input according to estimated probability. Advantages of DoT are as follows: (i) It ensures that the model learns using a more comprehensive input distribution
, which thus close performance gaps of using multi-view and single-view inputs in inference; (ii) different from random selection, using DoT will not degrade model's feature learning capability; (iii) it also narrow information gap between contrastive learning branch (single-view input) and generation branch (multi-view input).


The contributions of this work can be summarized as follows. (i) We identify the lack of multi-view input problem of existing medical report generation methods and propose a MvCo-DoT network for better medical report generation. (ii) A multi-view contrastive learning (MvCo) strategy is first proposed to utilize the multi-view information of chest X-Ray images for better model learning, while a domain transfer network is then proposed to ensure model can achieve good performances using only single-view inputs in inference stage.
(iii) Extensive experiments on a public dataset (IU X-Ray) show that: first, our proposed MvCo-DoT model greatly outperforms existing medical report generation baselines in all metrics; second, MvCo and DoT are both effective and essential for the model to achieve the superior performances; third, MvCo-DoT can achieve almost the same performance as multi-view inputs using only single-view inputs, which greatly saves the patients' time and money. 

\vspace{-0.8em}
\section{Methodology}
\vspace{-0.5em}

We propose a multi-view contrastive domain transfer network for medical report generation. As shown in Fig.~\ref {architecture}, we will adopt the architecture of generation branch and contrastive learning branch. Contrastive learning branch is used for inter-view mutual information mining, and a generation branch is used for report generation. The two processes are alternately performed during training.

\vspace{-1.1em}
\subsection{Multi-View Contrastive Learning}
\label{sec:subhead}
\vspace{-0.7em}

In order to mine and utilize the mutual information between medical images of different views, we propose semantic-based multi-view contrastive learning method with~\cite{xu2020reinforced} as the backbone network. Specifically, we concatenate the contextual semantic representations $c_f$, $c_l$ and hidden layer information $h_f$, $h_l$ decoded from different views and project them onto the same implicit space for comparison.

\vspace{-1em}
\begin{small}
\begin{equation} 
x_f=\psi(Concat(c_f,h_f)) , x_l=\psi(Concat(c_l,h_l)),
\end{equation}
\end{small}
\vspace{-1.5em}

\noindent where $c_f$ and $c_l$ are the context vectors of the last step of the LSTM in the twin M-Linear decoders, $h_f$ and $h_l$ are the corresponding hidden layer vectors. $\psi(\cdot)$ is modeled as two fully connected layers with ReLU activations, according to~\cite{nair2010rectified}. Then we maximize the semantic concordance between the frontal and lateral view of the same patient while minimizing the similarity between different patients. The multi-view contrastive loss function is defined as

\vspace{-1em}
\begin{small}
\begin{equation} 
L_{MvCo}=-\log\frac{exp(sim(x_l,x_f)/\tau_c)}{\sum_{k=1}^{2N}{\mathbbm{1}_{[k\ne{l}]} \ {exp(sim(x_l,x_k )/\tau_c)}}},
\end{equation}
\end{small}
\vspace{-0.5em}

\noindent where $sim(x_l,x_k)$ represents the cosine similarity,
$sim(x_l,x_k)$ $= {{x_l}^\top{x_k}}/{\parallel x_l \parallel \parallel x_k \parallel}$
and $\tau_c$ is temperature parameter.

\vspace{-0.8em}
\subsection{Domain Transfer Network
}
\label{ssec:subsubhead}
\vspace{-0.5em}

To overcome the domain shift problem caused by the input distribution gap between training and testing phases, We propose a domain transfer network based on adaptive input selection. First, we project the frontal and lateral visual features $f_{frontal}$ and $f_{lateral}$ extracted by ResNet101\cite{he2016deep} into two different latent spaces, to obtain more discriminative visual embeddings $F_{frontal}$ and $F_{lateral}$ for different views.

\vspace{-1em}
\begin{small}
\begin{equation} 
F_{frontal}=\phi_f(f_{frontal}) ,F_{lateral}=\phi_l(f_{lateral}),
\end{equation}
\end{small}
\vspace{-1.1em}

\begin{figure*}[!t]
\vspace{-2.5em}
\centering
\includegraphics[width=\textwidth]{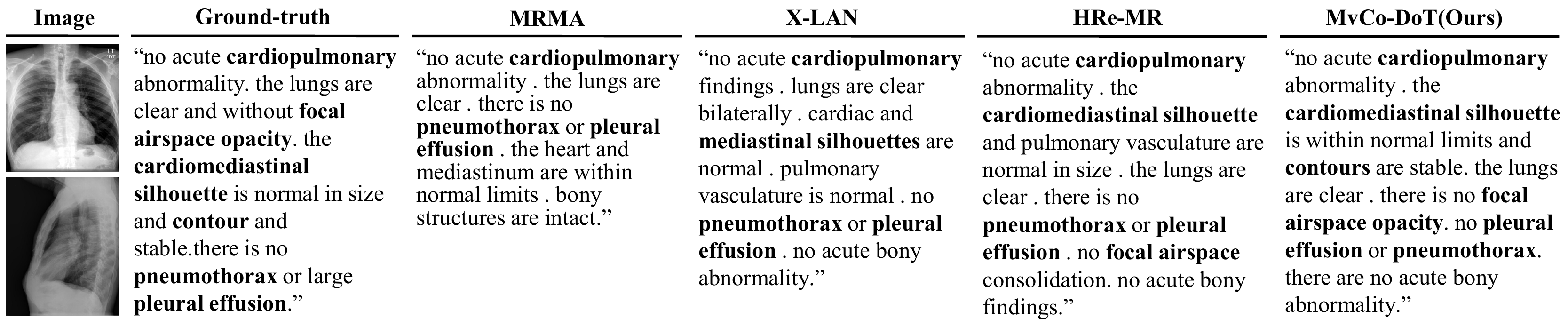} 
\vspace{-2em}
\caption{Example of reports generated by our MvCo-DoT model and beaslines.
}  
\vspace{-1.5em}
\label{result}    
\end{figure*}

\noindent where  $\phi_f(\cdot)$ and $\phi_l(\cdot)$ are modeled as fully connected layers with ELU activations. Afterward, we add the visual embeddings of these different views to obtain $F_{fusion}$, which replaces the operation of direct concatenation of original features commonly used in previous work. Then, in order to make the model get the most useful information input and better balance the use of frontal and lateral view information, we adaptively decide to input a single feature or mixed features through action sampling. The action space $A \in \mathbb{R}^{1\times3}$ is defined as $F_{frontal}$ (when $i=0$), $F_{lateral}$ (when $i=1$), or $F_{fusion}$ (when $i=2$).
\vspace{0.5em}


This non-deterministic approach enables the model to adaptively select optimal input to obtain the maximum amount of visual information for each image. To circumvent the technical problem that binary sampling actions cann't participate in backpropagation, we utilize random sampling based on \emph{Gumbel-Softmax} distribution. This reparameterization trick has been used in reinforcement learning to enable discrete decision ~\cite{jang2016categorical}. Non-differentiable action values are replaced by differentiable samples from \emph{Gumbel-Softmax} distribution.
Specifically, we concatenate the global visual features $F_{frontal}$ and $F_{lateral}$, which are multi-scale fusions of frontal and lateral views in the latent space and taken as a comprehensive information basis for the current action selection. It is sent to a linear layer through the fully connected layer to obtain the action confidence warehouse $P \in \mathbb{R}^{1\times3}$.

\vspace{1.5em}
\begin{table}[!t] 
	\caption{Results of MvCo-DoT and the SOTA baselines on IU X-Ray, where B, ME and RO stand for BLEU, METEOR and ROUGE-L, and all models are re-implemented by us. 
	}
	\centering   
	\small
	\vspace{0.5em}
	\begin{tabular}{l|p{0.6cm}<{\centering}p{0.6cm}<{\centering}p{0.6cm}<{\centering}p{0.6cm}<{\centering}p{0.6cm}<{\centering}p{0.6cm}<{\centering}p{0.6cm}<{\centering}}
		\hline  
		 Model & B-1 & B-2 & B-3 & B-4 & ME & RO  \\ \hline  
		 Top-down~\cite{anderson2018bottom} & 0.2822 & 0.1866 & 0.1241 & 0.0830 & 0.1455 & 0.3330    \\
		RTMIC ~\cite{10.1007/978-3-030-32692-0_77} & 0.3448 & 0.2188 & 0.1484 & 0.1063 & 0.1509 & 0.2890  \\
		MRMA~\cite{xue2018multimodal} & 0.3820 & 0.2520 & 0.1730 & 0.1200  &  0.1630 & 0.3090  \\ 
		X-LAN~\cite{pan2020x} & 0.3826 & 0.2724 & 0.1949 & 0.1405 & 0.1750 & 
		0.3441 \\
        R2Gen~\cite{chen2020generating}
  & 0.4349 & 0.2802 & 0.1868 & 0.1510 & 0.1773 & 0.3509 \\
		HRe-MR~\cite{xu2020reinforced} 
  & 0.4265 & 0.3025 & 0.2119 & 0.1502 & 0.1871 & 0.3608 \\
		\textbf{MvCo-DoT}  & \textbf{0.4533} & \textbf{0.3180} & \textbf{0.2228} & \textbf{0.1568} & \textbf{0.1958} & \textbf{0.3743}   \\
		\hline 
	\end{tabular} 
	\vspace*{-0.5em}
	\label{Table:baseline} 
\end{table}

\vspace{-2em}
\begin{table}[!t] 
\vspace{-1.5em}
	\caption{Results of ablation experiments on IU X-Ray. 
	}
	\centering   
	\small
	\vspace{0.5em}
	\begin{tabular}{l|p{0.6cm}<{\centering}p{0.6cm}<{\centering}p{0.6cm}<{\centering}p{0.6cm}<{\centering}p{0.6cm}<{\centering}p{0.6cm}<{\centering}p{0.6cm}<{\centering}}
		\hline  
		 Model & B-1 & B-2 & B-3 & B-4 & ME & RO  \\ \hline  
		 Base-Cat & 0.4175 & 0.2813 & 0.1951 & 0.1400 & 0.1820 & 0.3604 \\
		MvCo-Cat & 0.4373 & 0.3062 & 0.2139 & 0.1482 & 0.1933 & 0.3609  \\
		MvCo-Fusion  & 0.4440 & 0.3130 & 0.2196 & \textbf{0.1571}  & 0.1953   & 0.3698  \\
		\textbf{Ours}  & \textbf{0.4533} & \textbf{0.3180} & \textbf{0.2228} & 0.1568 & \textbf{0.1958} & \textbf{0.3743}   \\
		\hline 
	\end{tabular} 
	\vspace*{-1.75em}
	\label{Table:ablation} 
\end{table}
%

\vspace{-0.9em}
\begin{small}
\begin{equation} 
P=softmax(W_c(Concat(F_{frontal},F_{lateral}))),
\end{equation}
\end{small}
\vspace{-1.5em}

\noindent where $W_c$ represents the fully connected layer parameter matrix. Subsequently, the sampling module will generate action values $V \in \mathbb{R}^{1\times3}$, which defined as

\vspace{-1em}
\begin{small}
\begin{equation} 
\hspace{-0.5em}
V(a)=\frac {exp((\log(P_i(a)+g_i(a)/\tau_s)} {\sum_{j=1}^{3}{exp((\log(P_j(a)+g_j(a)/\tau_s)}},for\ i=1,2,3,
\end{equation}
\end{small}
\vspace{-0.8em}

\noindent where $g$ represents the noise sampled from standard \emph{Gumbel-Softmax} distribution, and $\tau_s$ is temperature parameter. The final input strategy is gained after $V$ through \emph{argmax} layer. During inference, $V$ is generated according to input directly. Sample action whose sample value in $A$ is calculated to be $1$, and reconstruct only the features corresponding to the action into final input feature $F_g$.

\vspace{-0.7em}
\begin{small}
\begin{equation} 
F_g= A(a_i ), V(a_i )=1,
\end{equation}
\end{small}

\vspace{-1em}
\section{Experiments}
\vspace{-0.5em}

\subsection{Experimetal Settings}
\label{ssec:subhead}
\vspace{-0.5em}
To evaluate the performance of our proposed MvCo-DoT, extensive experiments are conducted on public chest X-Ray image dataset IU X-Ray\cite{demner2016preparing}. We screen $3,111$ groups of cases from the dataset, each containing two X-Ray images of frontal and lateral views and a paired report. The dataset is randomly divided by $7$:$1$:$2$. In addition, words with frequency of less than $5$ are discarded and replaced with "UNK", and reported maximum generated length is set to $114$. We reimplement six state-of-the-art image captioning and medical report generation models as baselines, including Top-down~\cite{anderson2018bottom}, RTMIC~\cite{10.1007/978-3-030-32692-0_77}, MRMA~\cite{xue2018multimodal},X-LAN~\cite{pan2020x}, R2Gen~\cite{chen2020generating} and HRe-MR~\cite{xu2020reinforced}.We evaluated the models using six common automatic language generation metrics, including BLEU~\cite{papineni2002bleu}, METEOR~\cite{banerjee2005meteor}, and ROUGE-L~\cite{lin2004rouge}, where BLEU includes four n-gram-based metrics (BLEU-1 to BLEU-4 ).

We utilize ResNet-101 pre-trained on ImageNet~\cite{deng2009imagenet} to extract $2048$ dimensional region-level image features from the last convolutional layer. After being converted to visual embeddings of size $1024$, the encoder exploration with four stacks of M-linear attention blocks yields high-order synthetic features. During decoding process, we set size of hidden layer, word embedding dimension, and latent dimension of the projection layer to $1024$. During training, we first pre-train model with a batch size of $6$ for $60$ epochs using NVIDIA RTX 2080Ti GPUs, the model is optimized alternately by generation branch and contrastive learning branch. We set the base learning rate to $0.0001$, paired with a Norm decay strategy with $10,000$ warm-up steps, and used the ADAM~\cite{kingma2014adam} optimizer. We set $\tau_c$ to $0.1$ and $\tau_s$ to $0.3$. Finally, we train model with batch size of $2$ for $60$ epochs of reinforcement learning 
using beam search~\cite{vijayakumar2016diverse} with a beam size of $2$ to further improve model performance. We set the indicator-weighted mixed reward as our training reward~\cite{xu2020reinforced}, where weights of BLEU-1, BLEU-2, BLEU-3, BLEU-4, METERO, and ROUGE-L, are $2$, $2$, $1$, $1$, $2$, and $2$, respectively; and base learning rate is reduced to $0.00001$ and decayed by cosine annealing with a period of $15$ epochs.


\vspace{-1em}
\subsection{Main Results}
\vspace{-0.5em}
Table \ref{Table:baseline} shows experimental results of our proposed MvCo-DoT and five baselines on six natural language generation metrics, where all baselines are re-implemented. Furthermore, Fig.~\ref{result} presents some examples of generated reports .

In general, MvCo-DoT outperforms all state-of-the-art baselines among all metrics in Table \ref{Table:baseline}, because (i) our multi-view contrastive learning adequately performs multi-view mutual information learning to obtain superior performance, (ii) the same input for multi-view training and single-view testing is maintained, and task gap between contrastive learning branch and generation branch is narrowed, avoiding domain shift problem. Moreover, visualization results shown in Fig.~\ref{result} also support our findings, where MvCo-DoT generation obtains a more comprehensive and accurate report description than baselines. Thus, our proposed multi-view contrastive learning and domain transfer network are highly effective in enhancing the quality of report generation.

\begin{figure}[!t]
\vspace{-2em}
\begin{minipage}[a]{.48\linewidth}
  \centering
  \centerline{\includegraphics[width=3.5cm]{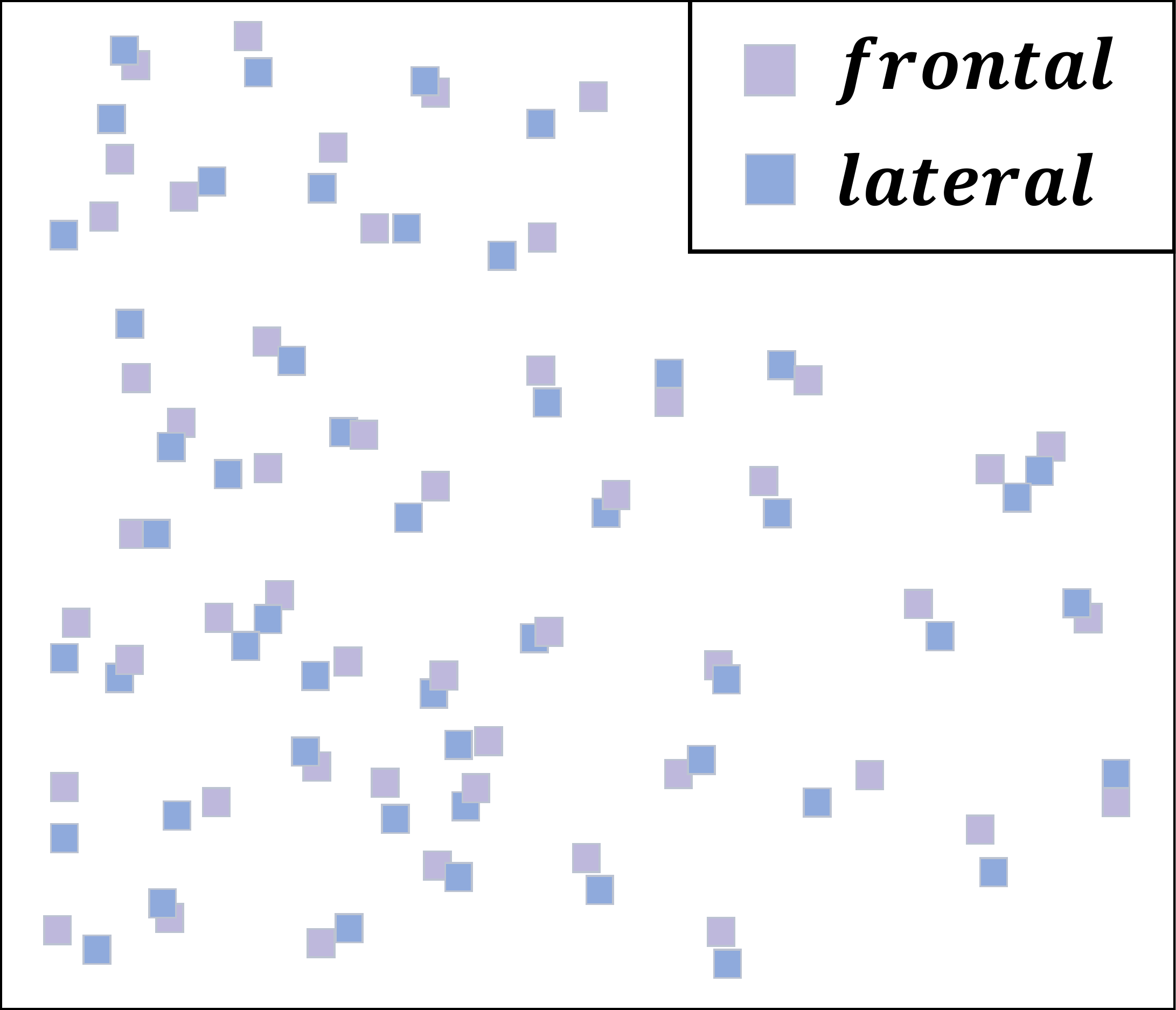}}
  \centerline{(a) Base}\medskip
\end{minipage}
\hfill
\begin{minipage}[a]{0.48\linewidth}
  \centering
  \centerline{\includegraphics[width=3.5cm]{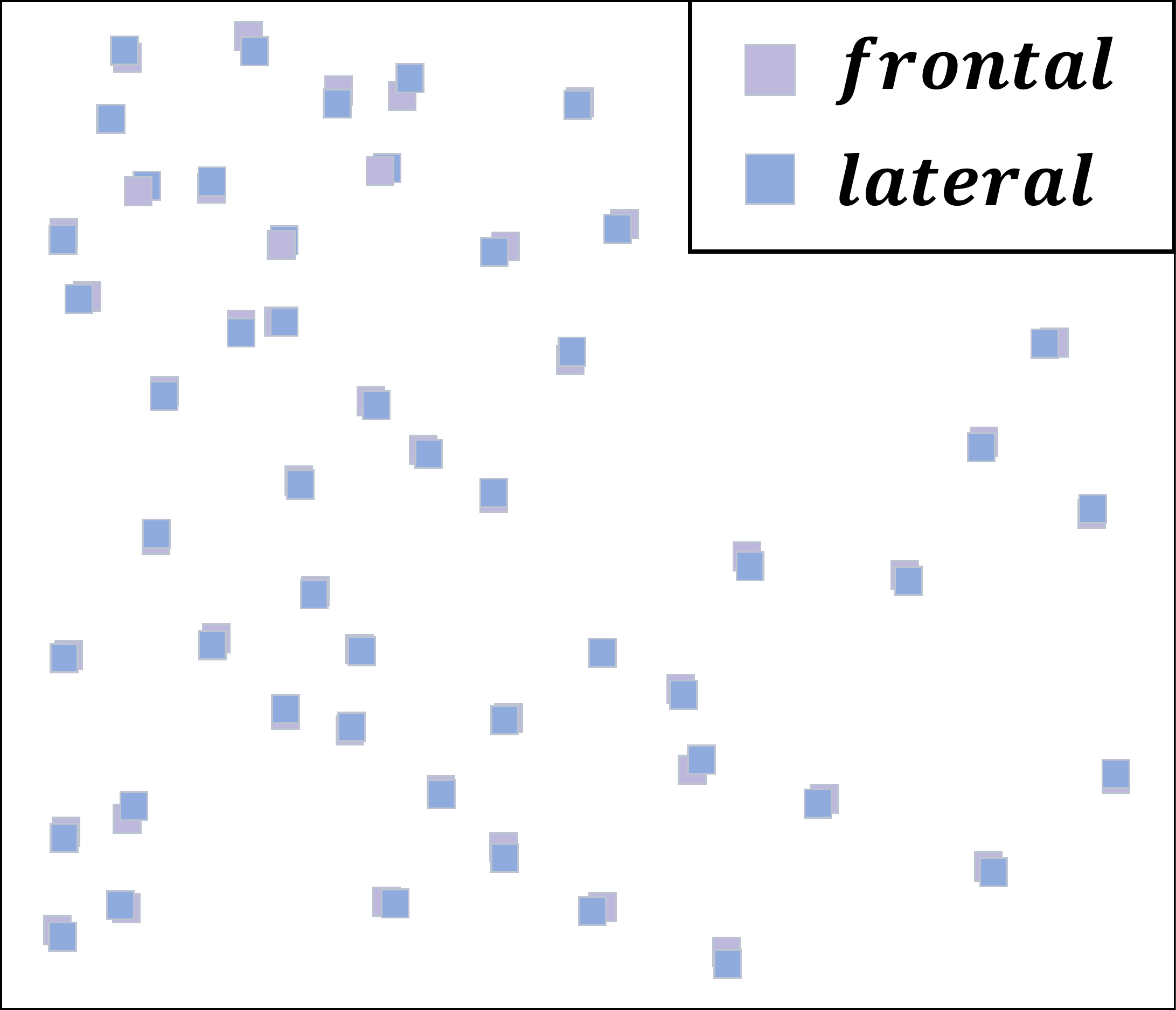}}
  \centerline{(b) MvCo}\medskip
\end{minipage}
\vspace{-1.25em}
\caption{Semantic feature embeddings in latent space.
}
\vspace{-1.7em}
\label{MvCo}
\end{figure}

\vspace{-1em}
\subsection{Ablation Study}
\vspace{-0.75em}

We further conduct a series of ablation experiments to demonstrate the effectiveness of each module of our proposed MvCo-DoT. We take generation branch as the base model, which utilizes raw feature concatenation of different views as input, called Base-Cat. Then we further implement two other versions of our MvCo-DoT: (i) Introduce a multi-view contrastive learning branch on base model, called MvCo-Cat, and (ii) using visual embedding fusion instead of original feature direct concatenation, called MvCo-Fusion. In Table \ref{Table:ablation}, we observe that all metrics of MvCo-Cat are superior higher than Base-Cat, as mutual information mined by multi-view contrastive learning helps to focus on salient lesions, explore deep semantic features and enable multimodal inference. The MvCo-Fusion score is further improved, indicating that feature fusion strategy is more suitable for our task. Next, we use \emph{Gumbel-Softmax}-based random sampling to adaptively select input strategy to obtain final model MvCo-DoT. Overcoming input distribution gap enables model to improve cross-domain transferability while reducing the task distance between the contrastive learning branch and generation branch, which makes it the highest score among all versions of the model. In conclusion, our proposed MvCo learning strategy and DoT network are very effective and essential to improve the accuracy of automatic generation of medical image reports.

\vspace{-1em}
\subsection{Additional Results}
\vspace{-0.75em}

\begin{figure}[!t]
\vspace{-2em}
\begin{minipage}[a]{.48\linewidth}
  \centering
  \centerline{\includegraphics[width=4.0cm]{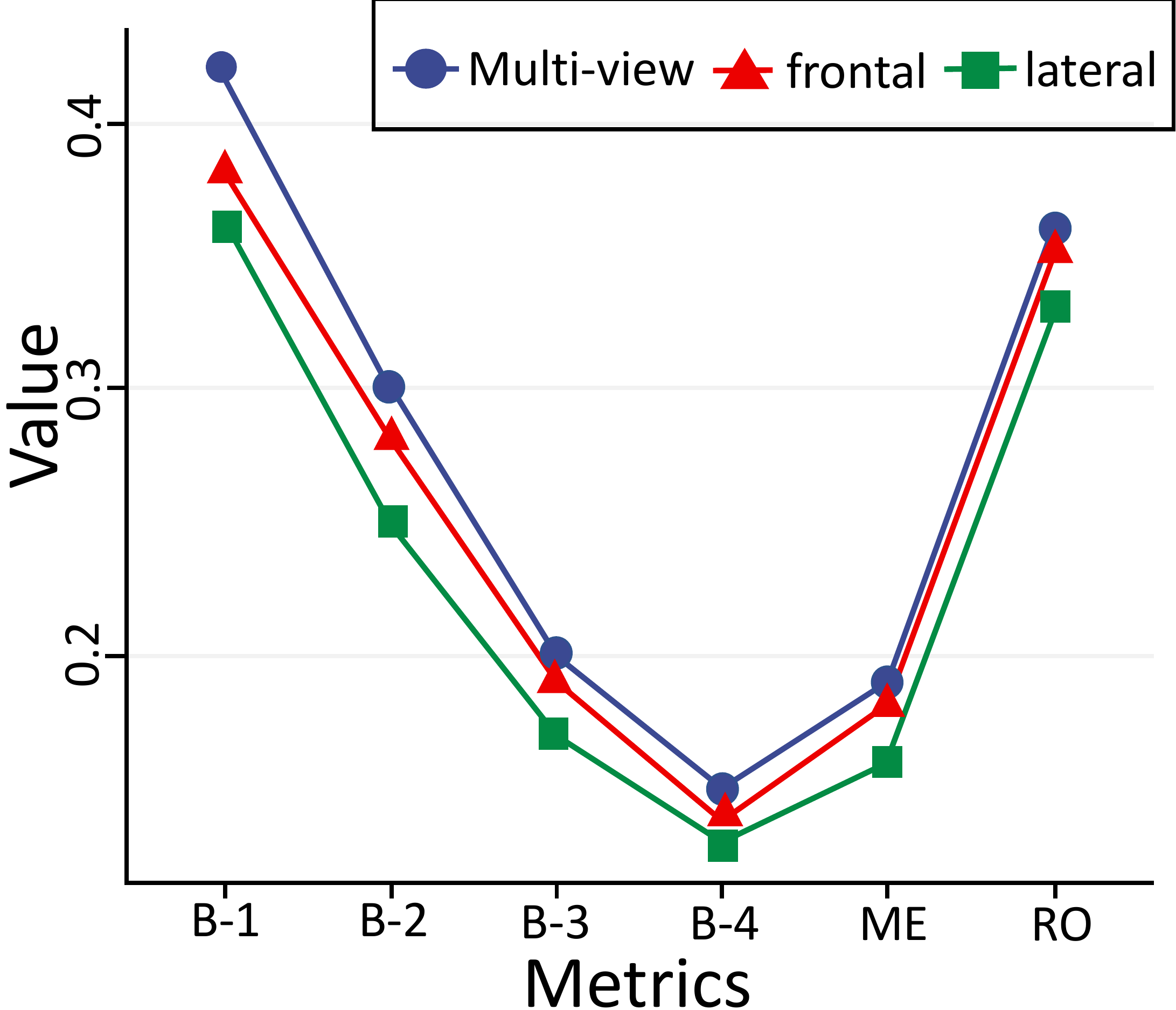}}
  \centerline{(a) MvCo}\medskip
\end{minipage}
\hfill
\begin{minipage}[a]{0.48\linewidth}
  \centering
  \centerline{\includegraphics[width=4.0cm]{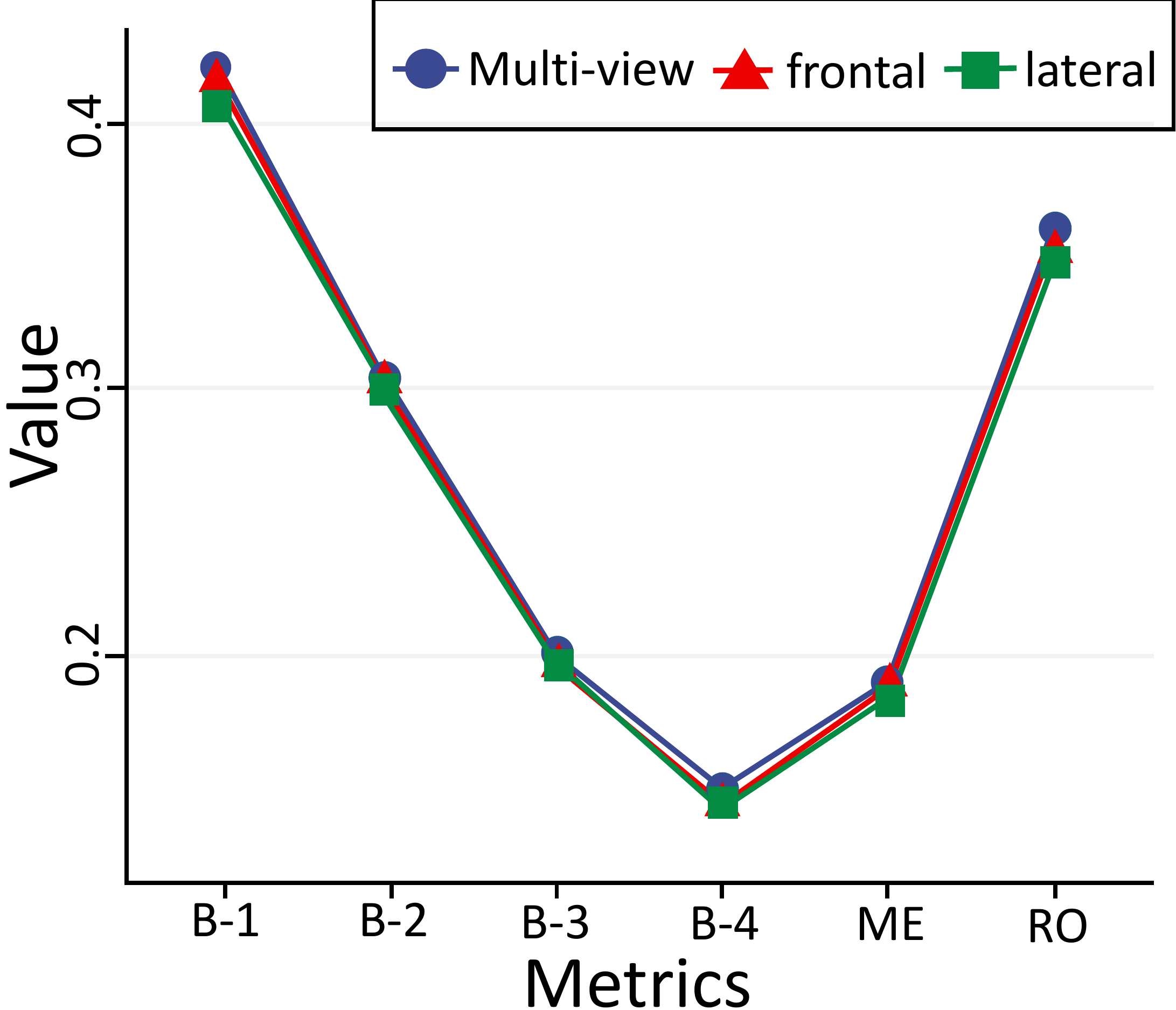}}
  \centerline{(b) MvCo-DoT}\medskip
\end{minipage}
\vspace{-1.2em}
\caption{Comparison of the performance of MvCo and MvCo-DoT for reporting inference using different inputs.}
\vspace{-1.7em}
\label{DoT}
\end{figure}

In this section, we investigate the impact of semantic-based multi-view contrastive learning on model performance and advantages of domain transfer networks based on adaptive inputs. Fig.~\ref{MvCo} shows the distance variation of frontal and lateral view semantic embeddings in the implicit space. Compared with generation branch Base in (a), multi-view contrastive learning MvCo in (b) can make the semantic embeddings of different views closer. This is because the model learns mutual information between different views, thereby decoding feature vectors with more semantic consistency. In addition, in Fig.~\ref{DoT}, compared with the pure multi-view comparison MvCo in (a), domain transfer network MvCo-DoT with adaptive input selection in (b) can generate high-scoring reports under any single view, which well solves domain shift caused by different input distributions during multi-view training and single-view testing. Moreover, the model performance is further improved compared to (a), which means that the same input distribution also reduces the gap. The model obtains the optimal representation of the two, promoting each other.

\begin{figure}[!t]
\vspace{1.1em}
\begin{minipage}[a]{.48\linewidth}
  \centering
  \centerline{\includegraphics[width=4.0cm]{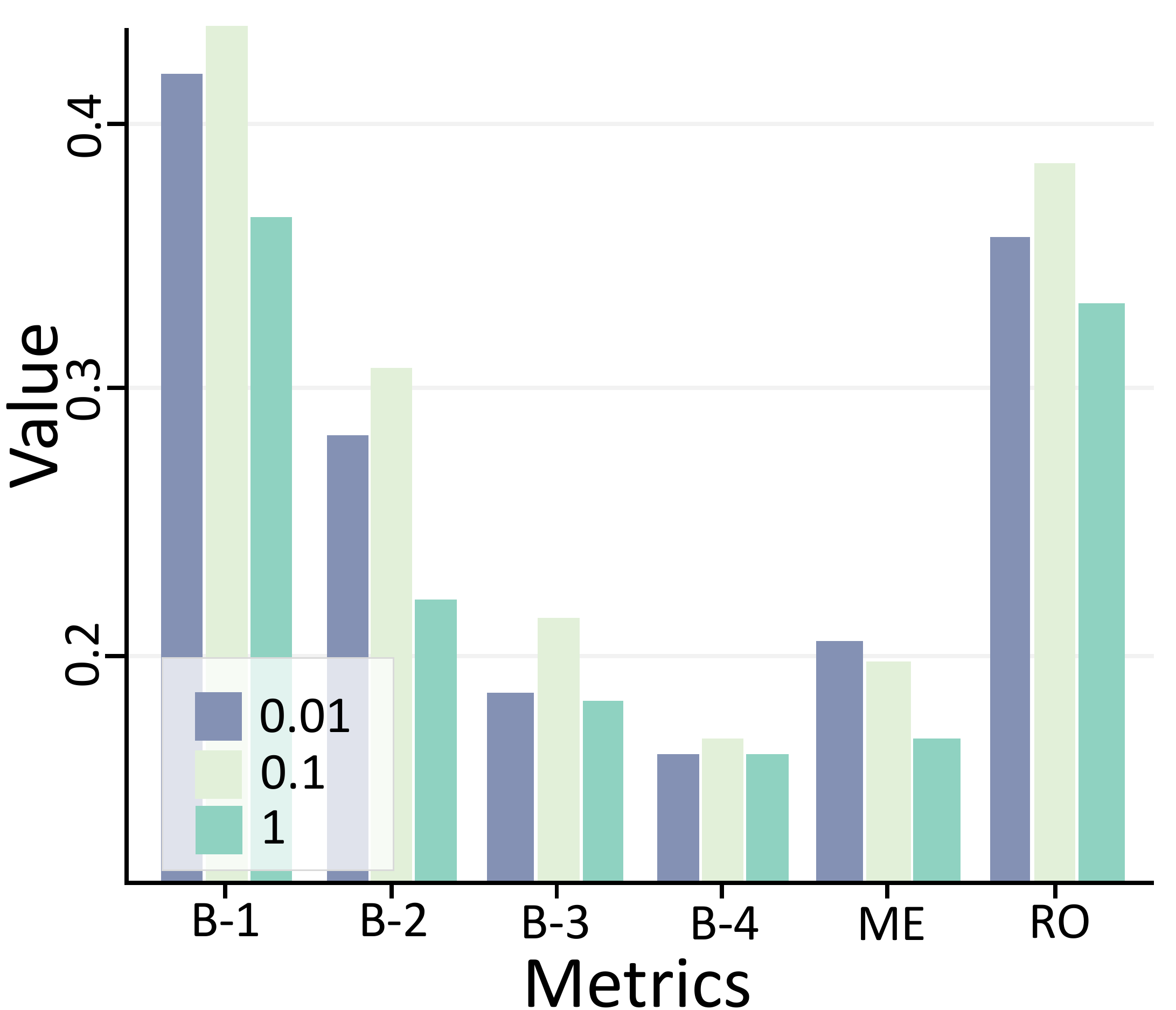}}
  \centerline{(a) different values of $\tau_c$}\medskip
\end{minipage}
\hfill
\begin{minipage}[a]{0.48\linewidth}
  \centering
  \centerline{\includegraphics[width=4.0cm]{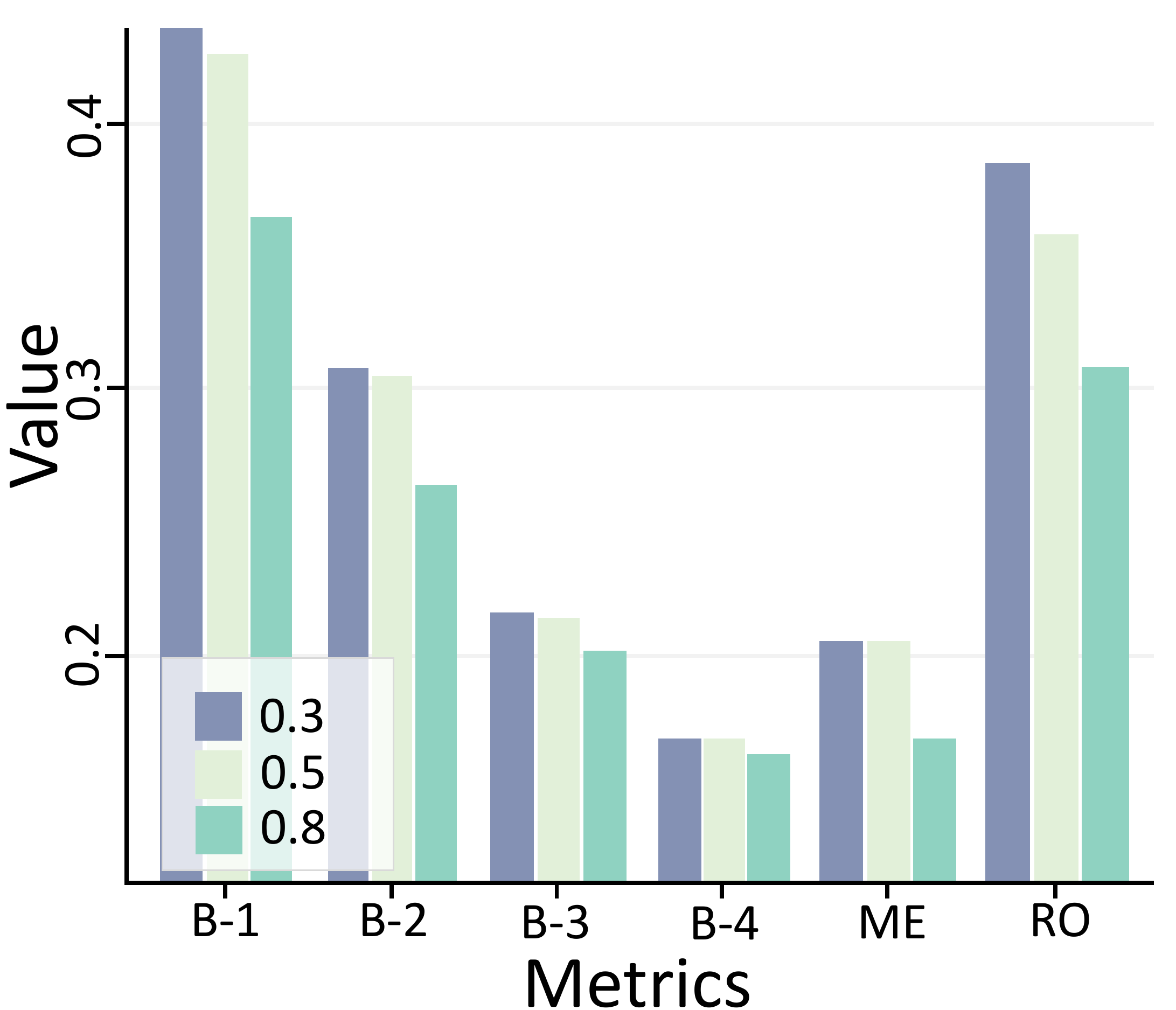}}
  \centerline{(b) different values of $\tau_s$}\medskip
\end{minipage}
\vspace{-1.2em}
\caption{Results of varying hyperparameters $\tau_c$ and$\tau_s$.}
\vspace{-1.2em}
\label{param}
\end{figure}

\vspace{-0.7em}
\subsection{Effect of Varying Hyper-Parameters $\tau_c$ and $\tau_s$}
\vspace{-0.5em}

$\tau_c$ and $\tau_s$ are the temperature parameter of the contrastive loss and \emph{Gumbel-softmax} distribution, respectively. In Fig.~\ref{param}, according to (a), in the range of $0.01$ to $1$, the performance of the model fluctuates with the size of $\tau_c$, and according to (b), smaller $\tau_s$ in the range of $0.3$ to $0.8$ is better. Therefore, we need certain tuning parameters to achieve the best performance. In this model, we set $\tau_c$ to $0.1$ and $\tau_s$ to $0.3$.

\vspace{-0,7em}
\section{Conclusions}
\label{sec:majhead}
\vspace{-0.7em}

In this paper, we proposed a multi-view contrastive domain transfer network (MvCo-DoT) for medical report generation. We mined mutual information between different views of chest X-Ray using multi-view contrastive learning based on semantic information to aid model learning. We also closed the input distribution gap between training and inference stages and contrastive learning branch and generation branch through domain transfer network based on adaptive input selection to address the domain shift problem. We performed extensive experiments on the publicly available dataset IU X-Ray, demonstrating the superiority and effectiveness of our proposed method. Future researches may include the usage of the proposed method in imbalanced learning task~\cite{wang2021rsg}, and incorporate more deep reinforcement learning techniques~\cite{DRLHomoPre2022} to enhance the deep generation model's learning capabilities.

\vspace{-0.65em}
\section{Acknowledgments}
\label{sec:majhead}
\vspace{-0.65em}

This work was supported by the National Natural Science Foundation of China under the grants 62276089, 61906063 and 62102265, by the Natural Science Foundation of Hebei Province, China, under the grant F2021202064, by the ``100 Talents Plan'' of Hebei Province, China, under the grant E2019050017, by the Open Research Fund from Guangdong Laboratory of Artificial Intelligence and Digital Economy (SZ) under the grant GML-KF-22-29, and by the Natural Science Foundation of Guangdong Province of China under the grant 2022A1515011474.

\vfill\pagebreak

\clearpage
\begin{spacing}{0.9}
  \bibliographystyle{IEEEbib}
  \small
    \bibliography{refs_bib.bib}
\end{spacing}
\end{document}